%% file: iclr2019_conference.tex
\documentclass{article} 
\usepackage{iclr2019_conference,times}

\input{math_commands.tex}

\usepackage{hyperref}
\usepackage{url}
\usepackage{graphicx}
\usepackage{subfig}
\usepackage{array}
\usepackage{hyperref}
\usepackage{sectsty}
\usepackage{titlecaps}

\title{Current Limitations of Language Models: What You Need is Retrieval
}

\author{Aran Komatsuzaki \\
Georgia Institute of Technology\\
EleutherAI\\
\texttt{akomatsuzaki3@gatech.edu} \\
}

\iclrfinalcopy

\begin{document}

\maketitle

\begin{abstract}
We classify and re-examine some of the current approaches to improve the performance-computes trade-off of language models, including (1) non-causal models (such as masked language models), (2) extension of batch length with efficient attention, (3) recurrence, (4) conditional computation and (5) retrieval. We identify some limitations (1) - (4) suffer from. For example, (1) currently struggles with open-ended text generation with the output loosely constrained by the input as well as performing general textual tasks like GPT-2/3 due to its need for a specific fine-tuning dataset. (2) and (3) do not improve the prediction of the first $\sim 10^3$ tokens. Scaling up a model size (e.g. efficiently with (4)) still results in poor performance scaling for some tasks. We argue (5) would resolve many of these limitations, and it can (a) reduce the amount of supervision and (b) efficiently extend the context over the entire training dataset and the entire past of the current sample. We speculate how to modify MARGE to perform unsupervised causal modeling that achieves (b) with the retriever jointly trained.
\end{abstract}

\section{Classification of Recent Language Model Approaches}
\label{secclass}
\begin{figure}
    \centering
    \subfloat{{\includegraphics[width=\linewidth]{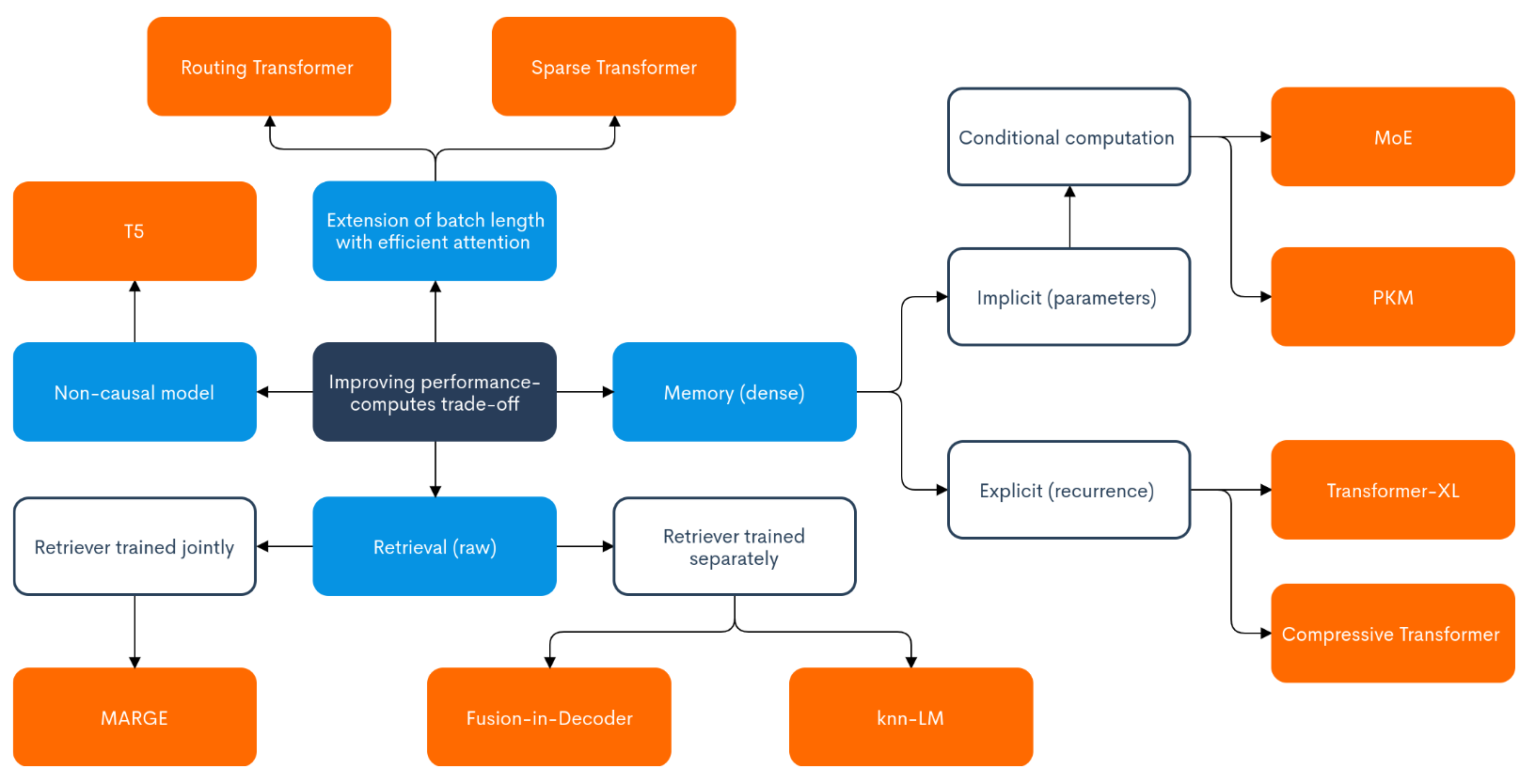} }}%
    \caption{Classification of recent language model approaches to improve performance-computes trade-off. Blue indicates a major category. Orange indicates an example. Batch length in this paper is defined to be the length of a minibatch across timestep dimension.}
    \label{classification}
\end{figure}

We consider some of the successful recent language model approaches to improve the performance-computes trade-off of language model \citep{kaplan} and classify them for the ease of our argument. Each category tends to be orthogonal to each other, so that they can often be used in combination for further gain. Fig. \ref{classification} summarizes our classification. There are four major categories: non-causal models, extension of batch length with efficient attention, memory and retrieval. We note that batch length in this paper is defined to be the length of a minibatch across timestep dimension. We avoid calling it context length to be more specific. In this paper, the default modality of our consideration is text unless specified otherwise. Also, memory and retriever are defined to possess a certain property addressed later in this section, so their usage may differ from the usual one. Efficient attention also refers to non-attention alternatives with better complexity, such as variants of convolution, for the convenience of our argument. 

Non-causal models in this paper are defined to be the model that predicts a token using future information. The example includes various BERT variants, including BART and T5 \citep{bert,t5,xlnet,roberta,longformer,bird,bart, unifiedqa}. On the other hand, causal models include the original Transformer encoder-decoder \citep{tra} and GPT-2/3. Though we usually consider the decoder-only model as the latter for a causal model, it should be clear from the context whether a causal model refers to the former, the latter or both. 

The most long-range language modeling approaches that rely on efficient attention without recurrence or memory, including Reformer \citep{reformer}, Routing Transformer \citep{rout}, Big Bird \citep{bird} and many others \citep{longformer, synthesizer, adapattn, sparse}, extend batch length with efficient attention without suffering from quadratic complexity in terms of length. Notably, the approaches as Transformer-XL \citep{xl} and Compressive Transformer \citep{comp} are not included in this category, since they are recurrent and their goal is to extend the context rather than batch length per se. 

It is reasonable to assume that simply extending batch length may not be enough to cover all the information needed for prediction. To extend the context beyond batch length, one needs to resort to either memory or retriever, with which the model can attend to stored relevant information without increased batch length. Essentially, these methods enable attention over the entire training dataset and the context of the current sample.  

In this paper, memory-based methods and retrieval-based methods refer to the models that are equipped with memory and retriever, respectively. Memory is defined to retrieve relevant information from its own dense tensor that encodes the training dataset and the context of the current sample, whereas retriever is defined to retrieve relevant segments of data from the training dataset and the context of the current sample in the original, raw format.  

Notably, memory is more straightforward, since the parameters of a neural network can be considered as a form of memory, as they are updated through backpropagation so as to incorporate the input information to improve their prediction in subsequent iterations. We refer to this type of memory as implicit memory and any method of improving the capacity to store the input information in this way beyond the baseline as an implicit memory approach. On the other hand, if a model combines the input information with the stored activations from the previous iteration to use it for the next iteration in the forward pass, the sent activations is called explicit memory. In other words, the memory used in recurrence at the level of TBPTT segment (e.g. Transformer-XL) is referred to as explicit memory. It is also possible to store the activations and keep them as trainable parameters, which can be called hybrid memory, and they share the same properties as both types. For example, explicit memory approach includes Transformer-XL, Compressive Transformer and many others \citep{xlnet,memory}, while implicit memory approach includes simply increasing the parameter count and conditional computation approach, such as Sparsely-gated MoE (MoE) \citep{moe, gshard}, Product Key Memory (PKM) \citep{pkm} and others \citep{universal,surprise}.    

On the other hand, it is less clear how to design a good retriever, since the data to be retrieved is stored in a raw format unlike for memory. The retrieval-based approach can be distinguished by whether the retriever and the language model are trained jointly (end-to-end differentiably) or not, where the former refers to models as MARGE, KIF \citep{kif}, REALM \citep{realm} and CRISS \citep{criss}, and the latter refers to the rest of retrieval-based approaches. If they are not trained jointly, the retriever is usually either a tf-idf variant such as BM25 \citep{bm25} (e.g. \citep{fly}), DPR \citep{dpr} (e.g. Fusion-in-Decoder \citep{fusion}, RAG \citep{rag}) or a pre-trained language model (e.g. knn-LM and BERT-knn \citep{bertknn}). While many of these models have not been applied for unsupervised causal modeling, we will pose a general idea to convert them for unsupervised causal modeling. 

For discussion about some relatively successful examples of attempts to improve the performance-computes trade-off outside of the categories mentioned, the reader can refer to Appendix \ref{examples}. 
\section{Non-Causal Models}
\label{causal}

For various tasks in NLP, non-causal models tend to show better performance-computes trade-off than causal models. For example, T5 outperforms the causal baseline not only in discriminative tasks but also generative tasks such as summarization and translation. BART achieves lower perplexity over the causal baseline on the task of fine-tuning to a conversational dataset. Furthermore, it was recently reported that XLNet achieves a superior performance to the GPT-3 of the corresponding size at Winogrande with few-shot learning \citep{xlnet_tweet}. UnifiedQA, a T5-like pre-trained model fine-tuned with a general QA dataset, is evaluated with few-shot learning examples a la GPT-3, and it performs nearly on par with about 30 times larger GPT-3 (that is not fine-tuned) \citep{unifiedqa}. 



However, the current approach of non-causal models suffer from some limitations. 

Firstly, many real-life tasks are open-ended text generation with the output loosely constrained by the input in a way similar to many of the prompt-output generation tasks a la GPT-2 and long-range unsupervised text generation. However, non-causal models are rarely evaluated on this sort of tasks. This practice unfairly favors a certain family of models, including masked language models, since they tend to perform poorly compared with the causal models. More details can be found in Appendix \ref{open_ended}. While non-causal models excel at reading/encoder-intensive tasks, they perform poorly on some writing/decoder-intensive tasks. There needs to be more effort spent on seamlessly combining causal and non-causal models to achieve the best of both worlds.

\begin{figure}
    \centering
    \subfloat{{\includegraphics[width=0.9\linewidth]{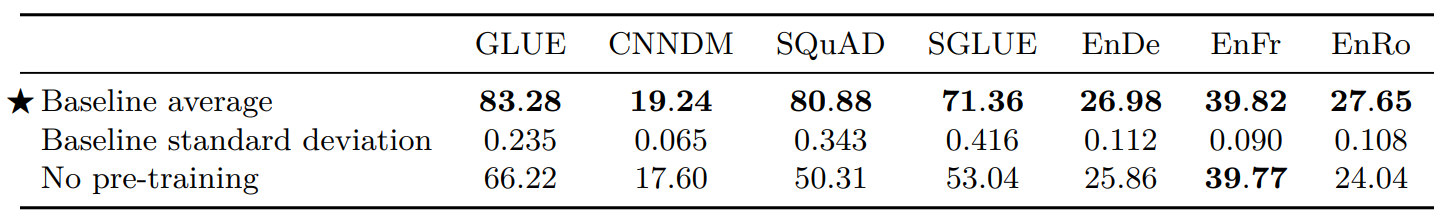} }}%
    \caption{\citep{t5}. The performance of the baseline T5 with and without fine-tuning on various datasets. Pre-training is beneficial for all fine-tuning datasets but WMT EnFr, which is the largest dataset.}
    \label{t5}
\end{figure}

Secondly, it is unclear how to fine-tune a non-causal model to perform an arbitrary textual task as GPT-2/3 at least as efficiently as GPT-2/3. Unlike GPT-2/3, non-causal models rely on the availability of a fine-tuning dataset relevant to a given task, which is usually not known beforehand. Hence, non-causal models are likely to perform poorly on the tasks where there is no fine-tuning dataset available for. This cannot be naively resolved by using a very general or large fine-tuning dataset as WebText to deal with various tasks, since the specificity of fine-tuning dataset is crucial for the power of fine-tuning. For example, \citet{t5} shows that conventionally fine-tuned T5 substantially outperforms the baseline fine-tuned with all the fine-tuning datasets combined, which implies that fine-tuning dataset needs to be specific. They also show that, when a fine-tuning dataset is large enough, the fine-tuned model degenerates into a causal model in terms of its performance, as the benefit of pre-training vanishes (Fig. \ref{t5}). Thus, there needs to be more investigation into how non-causal models would perform on previously unknown tasks compared with causal counterparts and how to improve the performance if necessary. 

\section{Extension of Batch Length With Efficient Attention}
\label{sec_ext}
Before beginning this section, we have a few caveats. Since the modality of our primary concern in this paper is text, our conclusion regarding the approach of extending batch length with efficient attention would differ if the modality of interest is different, especially if it has a structure more exploitable with an inductive bias such as image and video. In particular, we believe that, unlike text, such exploitable modalities will benefit substantially more from better efficient attention or related architectural modification than text. However, this is out of the scope of this paper.  

There have been many recent works that propose a language model that simply replaces full attention with an efficient attention to improve the computational complexity in hope of extending the batch length, i.e., from $O(Ld_{model}^2+L^2d_{model})$ to $O(Ld_{model}^2+f(L)d_{model})$, where $L$ is the batch length, and $f(L)$ is some function such that $f(L)=o(L^2)$. Notable example of $f(L)$ includes linear, log-linear and $O(L^{3/2})$. Since we have $Ld_{model}^2>L^{3/2}d_{model}$ if $L<d_{model}^2\sim 10^6$, in practice even $f(L)=O(L^{3/2})$ is sufficient to make the contribution of attention not to dominate that of linear layer. Thus, the computational complexity of many efficient attention models is already empirically on par with that of linear layers. In other words, one can no longer practically gain from improving the complexity of efficient attention any further.

\begin{figure}
    \centering
    \subfloat{{\includegraphics[width=0.5\linewidth]{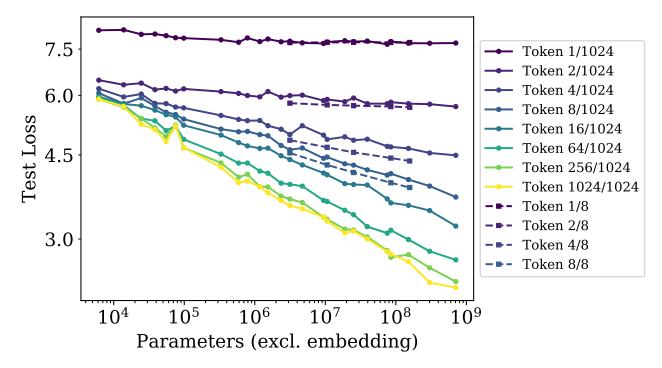} }}%
    \subfloat{{\includegraphics[width=0.5\linewidth]{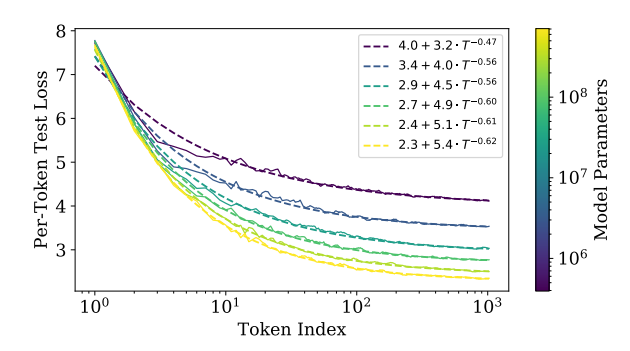} }}\\
    \caption{\citep{kaplan}. (Left): $n$-th per-token loss for various $n$ on models with various parameter counts and batch lengths. Training runs with shorter batch length $L = 8$ (dashed lines) perform better on early tokens, which implies that the model with longer batch length does not improve theIf anyone  loss of early tokens that have shorter context. (Right): $n$-th per-token loss for various $n$ on models with various parameter counts with $L = 1024$. Even for early tokens, increasing the parameter count results in a significant improvement in their loss.}
    \label{scaling}
\end{figure}

Most naturally available samples as well as the reasonable output of most tasks have rather limited length, though others (e.g. books) do not. For example, the average sample length of WebText is only about $1000$ tokens, and likewise for the webpages in Common Crawl. Also, when batch length is extended from $L$ (e.g. $1024$) to $L'$, this does not tend to result in substantial improvement in the loss of the first $L$ tokens of a sample. For example, Fig. \ref{scaling} (left) demonstrates this. On the other hand, other approaches such as increasing the number of parameters (implicit memory) would improve the loss for early tokens as shown in Fig. \ref{scaling} (right). It can be found on Appendix \ref{earlier} a more detail on how each approach improves or does not improve the loss of early (the first $L$) tokens of each sample. In particular, Appendix \ref{earlier} concludes that any approach that satisfies a certain condition (including retrieval and implicit memory approach) and improves the loss does improve the early token prediction. Since many text data have rather limited sample length, the gain from longer batch length due to efficient attention can be observed only on a small subset of entire text data. Even for the long samples, the prediction is poor for its early tokens, which results in poor generation at the beginning and cannot be mitigated by better loss of latter tokens.  

The approach of simply replacing full attention with efficient attention to extend the batch length has been successful to some extent. We summarize our observations about this approach in this section as well as that of Appendix \ref{details_ext} (including the one subsections not mentioned above) as follows:
\begin{enumerate}
    \item In practice, one cannot improve the effective computational complexity of efficient attention models any significantly further due to the contribution from linear layers.
    \item Appendix \ref{modelbatch} shows that improvement from extending the batch length would be limited at some point by the trade-off between model size and batch size that follows from \citet{kaplan}.
    \item This approach is unlikely to result in any improvement for modeling the first $L\sim 10^3$ tokens of a sample, i.e., the entirety of many samples of our interest, that tend to have short length, and the early tokens of long samples, whose suboptimal generation would also affect the generation of the later tokens. 
    \item Appendix \ref{extension_retrieval} argues that it is possible to combine efficient attention with retrieval-based approach to increase $k$ of $k$NNs retrieved for possibly further improvement of performance. However, it is uncertain if further improvement in efficient attention would bring a notable gain for this use over the existing efficient attention. 
 \end{enumerate}
\section{Memory-Based Approach}
\label{secmemory}
There are both pros and cons for each of explicit approach and implicit approach. Explicit memory (recurrence) is more sensitive to the recent past in contrast to implicit memory. For example, the perplexity of Transformer does not improve whether the minibatch of the previous iteration contains the immediate past of a sample in the current minibatch, while the performance of Transformer-XL (explicit memory) improves if its memory contains the activations of the immediate past of the current sample. This property is an advantage for explicit memory. However, explicit memory often cannot utilize information of the distant past such as the information from samples processed many iterations ago. While implicit memory (e.g. increasing the parameter count) can utilize such information by storing it in more parameters as can be seen from the fact that it improves the loss of early tokens (Fig. \ref{scaling}), this is not the case for existing explicit memory approaches such as Transformer-XL. The issue that recurrent models cannot utilize information well beyond the range of batch length through recurrence has been known for a long time, and there has been little progress in this direction. Thus, many of our conclusions for the approach of extending batch length with efficient attention apply to explicit memory as well. 

One may counter the above argument by noting that self-attention is known to perform poorly on some tasks recurrence excel at. We believe this is not a valid counter-argument as addressed in Appendix \ref{parity}.  

Implicit memory approach, by definition, aims for compressing as much information as possible in the parameters in a way such that it can retrieve them efficiently. Since having more parameters allows storing more information, conditional computation is more advantageous. The most successful conditional computation approach thus far is presumably MoE. Both MoE and PKM essentially expand the effective size of $d_{ff}$ and sparsely accesses to each unit as discussed in Appendix \ref{moepkm}. Obviously, there are other dimensions of Transformer that can be expanded with conditional computation, including depth and heads. We can also regard, for example, Routing Transformer as a conditional computation approach to let each head to attend to some timesteps only, though we separate them in this paper for the ease of argument. Though we have tried to conditionally expand depth and the heads, it has not saved performance-computes trade-off thus far. There are many works that have tried to expand depth conditionally \citep{universal,depthadaptive}, but there has been no such work that demonstrated a robust scaling and a substantial improvement in performance-computes trade-off. We hope this problem will be resolved soon. 

Since conditional computation aims for saving the computes and adding the parameters, the parameter GPU memory may eventually become a bottleneck if the model size keeps increasing. However, L2L proposed in \citet{constmem} reduces the parameter GPU memory from $O(L)$ to $O(1)$ by saving the parameters in CPU, where $L$ is the number of layers, while maintaining comparable speed. 

Despite the success of conditional computation, the models that are made larger and more capacious with conditional computation do not necessarily outperform retrieval-based methods, as argued in Appendix \ref{lim_cc}. This motivates for the use of retrieval-based approach in conjunction to conditional computation. 

This section can be summarized as follows:
\begin{enumerate}
    \item Explicit memory approach suffers from many of the same problems as extending batch length with efficient attention. Hence, most of our conclusions in the previous section applies to this method as well.
    \item Conditional computation, as a branch of implicit memory approach, brings a dramatic improvement of performance-computes trade-off, usually by efficiently approximating a Transformer with larger hyperparameters such as $d_{ff}$.
    \item Given that the vanilla models that spend more computes perform poorly at some tasks compared with substantially smaller retrieval-based models, retrieval-based approach should also be needed, though there still is a space left for conditional computation to bring more gains.
\end{enumerate}

\section{Retrieval-Based Approach}
\label{secretr}
\subsection{Retrieval as Self-supervision}
The supervision as few-shot learning and fine-tuning may not be always possible in the general real-life tasks, since there are numerous, diverse tasks, and naively finding a set of relevant samples for each task accurately may not be feasible. This necessitates the use of self-supervision to reduce the required amount of supervision.

The conventional method of construction of a dataset or feeding samples into a model can be thought of as a form of retrieval. Retrieval with a language model would partially automate this process and reduce the amount of supervision required to solve a given task. For example, CRISS, a retrieval-based model and the current state-of-the-art of unsupervised machine translation (even without back-translation) over various language directions with substantial improvement over its baseline, retrieves the target for a given input sequence. The retriever essentially is trained to learn what humans would feed for a given input sequence. Due to the ubiquity of this procedure in machine learning, retrieval should be a crucial component.  

\subsection{Retrieval Resolves Many Limitations Other Approaches Suffer From} 
As stated at the end of the previous section and Appendix \ref{lim_cc}, retrieval-based models, such as Fusion-in-Decoder and MARGE, often substantially outperform the Transformer language models that either have a larger model size or spend more computes on various tasks, which implies that it is not only sufficient to scale up a model but also equip it with a retriever to improve the performance of language model. Hence, retrieval complements conditional computation in some tasks. Retrieval also improves early token prediction as argued in Appendix \ref{earlier}. Retrieval may also help non-causal models perform general textual task through self-supervision.

As argued previously, efficient attention and recurrence have limited range of context both within the same sample and outside of it. However, retrieval can achieve indefinite context length within the same sample and the entire training dataset. For example, knn-LM achieves the state-of-the-art in Wikitext-103 for the amount of computes thanks to its context over the entire training dataset enabled by $k$NN search with faiss. While there has been no demonstration of achieving indefinite context length within the same sample with retrieval, in principle this can be similarly done as suggested later. 

Retrieval can also self-supervise few-shot learning of GPT-3 by feeding retrieved relevant samples. In fact, our proposed modified MARGE presented below trains the retriever that retrieves few-shot learning examples with the language model jointly, so that both components can benefit from the improvement of each other.  



\subsection{MARGE and Its Significance}
For retrieval-based models, it is more desirable to train the retriever and the language model jointly in a way such that the objective of the retriever is to maximize the performance of the language model, since models with separately trained retriever, such as knn-LM. There are various models with jointly trained retriever, but we focus on MARGE below, as MARGE also incorporates the trait of Fusion-in-Decoder, the attention bias for finding the similarity between source segments and target segments efficiently. MARGE performs respectably, for example, even without fine-tuning or back translation on unsupervised zero-shot machine translation with relatively small computes spent, not to mention that it also achieves the state-of-the-art performance on some tasks on translation and summarization with smaller computes spent. 

MARGE has the added attention bias set in a way such that it becomes larger between a source segment and a target segment closer to each other in terms of cosine similarity of their embeddings, and it computes the prediction loss of the target segments that are conditioned on the source segments. This design allows gradient update to make the embedder to embed a target segment and a source segment closer in terms of cosine similarity if the perplexity of the target segment conditioned on the source segment is lower. Thus, the model can find the source segments that give a lower perplexity to a given target segment, and simultaneously the language model conditioned on retrieved contexts can be trained. After many iterations, the clusters are no longer up-to-date to the trained embedder. Hence, the embedder is used to find more up-to-date clusters of related segments using $k$NN search of the newly computed embeddings.  



We outline the notable aspects of MARGE, while skipping the details irrelevant to our subsequent discussion:
\begin{itemize}
    \item We first divide each sample from the dataset into segments, each of which consists of $N$ consecutive tokens, where $N=512$ in the paper, and treat them as samples.
    \item We heuristically create a set of shards of segments, so that each segment is closely related to others in the same shard.
    \item MARGE uses the conventional Transformer encoder-decoder architecture with the usual prediction loss on the decoder. The normalized output of the first half layers of the encoder at the first timestep is used as the embedding of each segment for weight sharing.
    \item After each $I$ iterations, where $I=10000$ in the paper, we compute the embedding of each segment and the similarity between each embedding within each shard and sample clusters of $k$-nearest neighbors for batch construction. 
    \item Each batch consists of sub-batches, each of which consists of the same number of evidence (source) segments and target segments, which are similar to each other within the same shard and processed by the encoder and the decoder, respectively. 
    \item The encoder-decoder attention comes from every source segment to every target segment within the same sub-batch. The encoder-decoder attention from the source segment $z_i$ to the target segment $x_j$ is modified in a way such that $q\cdot k\to q\cdot k+\beta e(x_j)\cdot e(z_i)$ for a query $q$, a key $k$, the embedder $e$ and a trainable scalar $\beta$ with softmax taken over all the tokens in the source segments within the same sub-batch.
\end{itemize}

\subsection{Modified MARGE}
\begin{figure}
    \centering
    \subfloat{{\includegraphics[width=\linewidth]{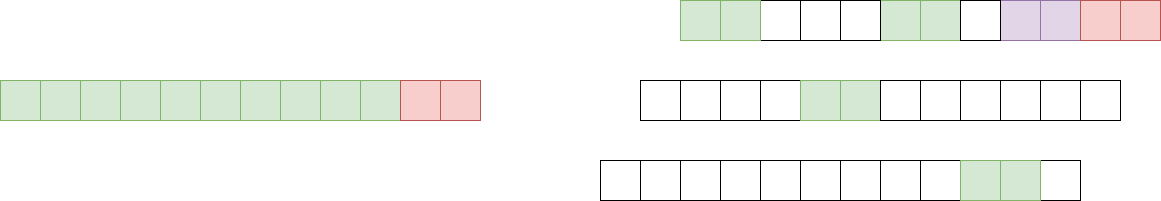} }}
    \caption{The conventional unsupervised causal modeling (left) vs. our modified MARGE
    (right). Note that each block consists of multiple tokens. If a model tries to predict the tokens in last two blocks (red), the attention is paid to the past tokens in the red region as well as the tokens in context (green). The modified MARGE reads the immediate context (purple) to find the relevant parts across the dataset and the past of the current sample in order to use them as a context.}
    \label{comparison}
\end{figure}

For combining retrieval with unsupervised causal modeling, the general idea is that, as in \citet{fly}, the current context is used (embedded in this case) to retrieve similar segments across the dataset and the past of the current sample, which are added to the context for the prediction. Unlike \citet{fly}, we would like to retrieve the relevant contexts multiple times per sample so as to retrieve the contexts more relevant to the most recently available context. Hence, when we divide a sample into multiple segments, each of which has length $N$, (i.e., a sequence $S_0,S_1,\ldots S_t,\ldots$, where $S_t$ is the $t$-th segment of the sample in the order of timestep), we want to make the value of $N$ reasonably small, e.g., $N\ll 512$, but large enough so that each segment makes sense by itself. If $S_t$ is a source segment, $S_{t+1}$ is used as its target segment. Using faiss, this allows unsupervised causal modeling over the entire training dataset as well as the entire past of the current sample. Fig. \ref{comparison} compares the conventional unsupervised causal modeling and our modified MARGE.  

While the original MARGE uses a heuristics exploiting the metadata of each document to collect similar segments into a shard, for more general case without metadata or to find similarity among samples with not necessarily similar metadata, the use of $k$NN search with faiss \citep{faiss} is preferred. At the beginning of the training, we can let each sub-batch to consist of the segments from the same sample (or from similar samples), since poorly initialized clusters are known to result in slow improvement. The model also needs to mask the encoder-decoder attention from a segment to its past segment in the same sample in order to preserve causality.  
\subsubsection*{Acknowledgments}
I wish to record my deep sense of gratitude and profound thanks to Loren Lugosch, Madison May, Lukasz Kaiser and Phillip Wang for their helpful, inspiring, well-thought feedback. 
\bibliography{iclr2019_conference}
\bibliographystyle{iclr2019_conference}
\section{Appendix}

\subsection{Some Other Approaches}
\label{examples}
\subsubsection{Learning to Summarize from Human Feedback
 \citep{feedback}}
\label{feedback}
This work achieves super-human-level summarization on TL;DR dataset by training a reward function on human feedback and fine-tuning a pre-trained causal model with PPO. This approach is very general for being applicable to any rank-able tasks with human evaluation easily available. Since modeling human-generated contents with MLE may be merely an imitation of human behavior and may not allow super-human performance, it is possible that this approach is more efficient for achieving super-human performance than MLE modeling. However, due to its need for a large amount of supervision, we do not investigate this approach further in our paper.       

\subsubsection{Progressive Generation of Long Text \citep{proggen}}
Progressive Generation of Long Text achieves substantially better text generation (human-evaluated) quality and diversity than GPT-2 by pre-training multiple Seq2Seq models, each of which "super-resolutes" a text, ultimately from a set of keywords into a complete text. However, since the evaluation is based on fine-tuning to a domain-specific dataset, the same remark as non-causal models apply. There needs more investigation into how it would cope with general textual tasks.  

\subsubsection{DeLighT: Very Deep and Light-weight Transformer \citep{delight}}
DeLighT, a Transformer language model, each of whose layer consists of a smaller feedforward network, a single head self-attention and DExTra, a module consisting of many layers of thin group linear layers, performs on par with the baseline Transformer language model with twice more parameters. Since group linear layer is slow without a custom CUDA kernel, the model would become faster than the baseline for a given performance if there is a custom CUDA kernel implemented for DExTra. However, since we often observe that a deep thin model is slower than a shallow wide model with the same parameter count using GPUs regardless of using a custom CUDA kernel, we are not certain if this would improve the performance-computes trade-off. Furthermore, since the source of improvement seems to stem from the use of DExTra, a block that has no interaction between different timesteps, we can assume that this module augments the feedforward network likely without any effect on the attention layer. Hence, DExTra is in a direct competition with MoE. While it is likely that they can be combined for further gain, the attraction of DExTra may diminish if MoE is also taken into the consideration.

\subsection{More Details of Sec. \ref{causal}: Non-causal Models}
\label{appendix_causal}
\subsubsection{Open-ended text generation with the output loosely constrained by the input}
\label{open_ended}
Let us consider the task of open-ended text generation in the sense that the input provided (e.g. prompt) is short and not very informative, while the expected output is substantially longer and more informative, in a way similar to how GPT-2 generates an article given an input. Since we expect a model to have the same autonomy as humans, the model has to be able to generate without detailed specifications from humans. This is in contrast to many of the tasks non-causal models are evaluated on (BART as a notable exception), which are often either discriminative tasks or generative tasks whose input sequence contains substantial information about the output, such as summarization and translation. In fact, \citet{bart} shows that BART and BERT variants perform poorly on none but the causal generative modeling of ELI5 dataset compared with the causal model, as they argue that the task on this dataset belongs to the aforementioned family of tasks.

\subsection{More Details of Sec. \ref{sec_ext}: Extension of Batch Length With Efficient Attention}
\label{details_ext}
\sectionfont{\titlecap}
\subsubsection{Improvement from extending the batch length would be limited at some point by the trade-off between model size and batch size}
\label{modelbatch}
According to \citet{kaplan}, the compute-optimal training strategy of causal Transformer language model is to allocate most of the increase in available computes ($C$) into parameter count ($N$) rather than batch size ($B$) or the number of iterations ($S$): \[N\propto C^{0.73},\quad B\propto C^{0.24},\quad S\propto C^{0.03}.\] Note that we have $B=bL$, where $b$ is the number of sequences per minibatch. In practice, $b$ has to be sufficiently large; otherwise, it is often observed that the resulting perplexity degrades due to lack of diversity in minibatch. Since optimal size of $B$ for even the largest model explored in \citet{kaplan} is about several millions of tokens, it is reasonable to assume that $L=\frac{B}{b}$ cannot be made arbitrarily large, possibly at most the order of million. Thus, it is safe to argue that an improvement from extending the batch length would be limited by the trade-off between model size and batch size. 

However, a caveat follows. The aforementioned compute-optimal strategy is under the assumption that the batch length is fixed ($L=1024$) and that WebText is used as the dataset. If $L$ can be enlarged and if a dataset with longer average sample size is used, the scaling exponents for compute-optimal training should change, though given the importance of batch size to the performance, it is reasonable to assume that the change is small. 

\subsubsection{Why does extension of batch length with efficient attention not improve the early token prediction?}
\label{inter_sample_extension}
To our assertion that extension of batch length does not improve the early token prediction, there is a good counter-argument: the loss of early tokens may be substantially improved if batch length is extended from $L\sim 1024$ to $L'\gg L$, so that numerous samples are covered and the model will utilize inter-sample information a la implicit memory approach and retrieval-based approach. As addressed before, it is very likely that $L'$ is bounded above (possibly at the order of million). Hence, if samples are ordered randomly, the samples in the same context are far from the $k$NNs from each other. Therefore, the performance gain from inter-sample information is likely to be negligible compared with the gain with retrieval-based approach, which can find the $k$-nearest neighbors. 
\subsubsection{Combination of retrieval-based approach with the approach of extending batch length with efficient attention}
\label{extension_retrieval}
It is possible to combine efficient attention with retrieval-based approach in order to increase $k$ of $k$NN samples to be used for prediction with a retriever, since a larger $k$ notably results in a better performance to some extent according to \citet{fusion}. However, there are two caveats. 

Firstly, the existing efficient attention may be already efficient enough that any further improvement of the method may not be beneficial for this use, as the empirical complexity of efficient attention would not be improved further according to Sec. \ref{sec_ext}. 

Secondly, the design of such efficient attention may depend on how the retrieval-based model is designed. For example, MARGE embeds each sample, so that the samples with the closest embeddings can attend to each other through attention. In this case, instead of applying a method as Routing Transformer, one can simply approximate the full attention with top-$k'$ approximation at sample-level for some reasonable $k'$, which does not require any sophisticated efficient attention method.

\subsubsection{How to assess if a given method improves early token prediction}
\label{earlier}
In this paper, we define "the improvement of early token prediction" to be the improvement of the loss of the first $L$ tokens of each sample for a small enough $L$. We set $L\sim 1024$, since the average length of a sample in the currently used gigantic text datasets is about $1000$ (for being a subset of Common Crawl) and since setting $L\sim d_{model}$ is a reasonable choice for making the full-attention computes comparable to that of the linear layer (as $d_{model}\sim 1024$).  

The most straightforward method to verify if a given model improves early token prediction or not is to compare the $n$-th per-token loss of early tokens for various $n$, which we have performed on various models, which led to our conclusion. However, we can also demonstrate to some extent that a given method improves the early token prediction without comparing the per-token loss for some cases. Models without recurrence tend to divide each sample into segments, each of which has length $L$, and they tend to be evaluated on a segment without being conditioned on the past tokens outside of the segment within the same sample. In this case, each segment is essentially treated as a sample for the lack of dependence on each other. Hence, the usual loss of the test dataset is equal to the loss of the first $L$ tokens of each sample. In other words, for a model without recurrence, if the segment length of the model is equal to that of its baseline, it suffices to compare the loss of the model with that of the baseline to see if the model improves the early token prediction, since "early token" cont. 

For example, knn-LM and its baseline are the same vanilla Transformer, except that the former is also conditioned on the distribution of the relevant tokens from the training dataset. The computes spent for the former is only marginally larger than that of the latter, while the improvement in the loss is substantial; hence, knn-LM improves the early token prediction for the given computes. A similar conclusion can be made for the GPT-2 with a retriever as in \citet{fly}, which is an augmented GPT-2 fine-tuned to predict conditioned on the sentences fed by a retriever. Since the fine-tuning computes are marginal relative to that of the pre-training and since the improvement in performance is large enough, we can argue that they too improve early token prediction for the given computes. As discussed later, many implicit memory models are the same as its baseline except for being an efficient approximation to a Transformer with a larger model size. Hence, this implies that such implicit memory models that improve the performance-computes trade-off, including MoE and PKM, improve the early token prediction for the given computes.    
\subsection{More Details of Sec. \ref{secmemory}: Memory-Based Approach}
\label{appmemory}
\subsubsection{The tasks recurrence excels at and self-attention struggles with}
\label{parity}
Self-attention is known to perform poorly on some tasks recurrence excel at. For example, \citet{theoretical} theoretically shows the former struggles with predicting the parity of a given long bitstring the latter excels at. However, TBPTT-segment-wise recurrence as explicit memory approach with self-attention usually needs to have long enough batch length (e.g. $128$ tokens) to make its performance-computes trade-off reasonably high on various tasks of practical interest. With this batch length, the local behavior of explicit memory model should be identical to that of self-attention. Hence, it should struggle just as self-attention. Furthermore, the tasks self-attention struggles with and recurrence excels at tend to be difficult even for humans. For example, if humans, without provided any instruction, are forced to find a pattern in an extremely long bitstring to figure out that the problem asks for the parity, they too would struggle with the task.
\subsubsection{More details about MoE and PKM}
\label{moepkm}
Let us consider a feedforward layer with a very large $d_{ff}$. MoE can imitate this by dividing this layer into many feedforward layers with smaller $d_{ff}$, each of which is an expert of MoE, and the gating of MoE controls which expert to be activated for a given input vector. On the other hand, PKM can imitate a large feedforward network by exploiting the following structural sparsity of feedforward layer we have observed: masking every units but the top-$k$ units of $d_{ff}$ units by zero does not affect the training of the model substantially even if $k\ll d_{model}$ (we observed that setting $k=\frac{d_{ff}}{10}$ does not affect the perplexity at all in some cases). Hence, PKM tries to predict which units are the top-$k$ units for a given input vector by its product key mechanism and efficiently computes the output by fused gather-matmul, which in practice results in slower processing than MoE due to memory bottleneck.   

Scaling up the model size by evenly increasing $d_{ff}$, $d_{model}$, $h$ (the number of heads) and depth, without increasing $d_k$ (the dimension of each head) tends to beat other approaches. This is presumably why GPT-2 and GPT-3 were scaled up in a similar manner. Let us call this optimal model scaling. Hence, increasing $d_{ff}$ only for scaling is suboptimal, and it is likely that the scaling law using this scaling with respect to the parameter count may not hold as long as that of the optimal scaling, not to mention that the power law exponent is likely worse. However, the gain from MoE and PKM is still substantial.

\subsubsection{Limitations of conditional computation}
\label{lim_cc}
\begin{figure}
    \centering
    \subfloat{{\includegraphics[width=0.9\linewidth]{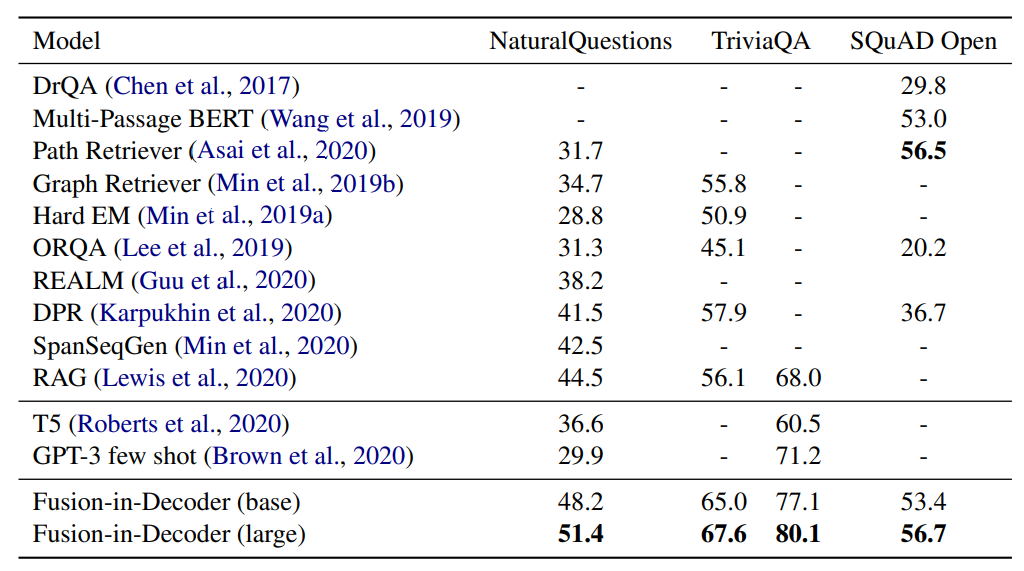} }}
    \caption{\citep{fusion}. Comparison of the state-of-the-art QA models.}
    \label{fid}
\end{figure}

More successful conditional computation methods, such as MoE, are mostly an efficient approximation to a Transformer architecture with larger model hyperparameters, such as $d_{ff}$. Hence, its performance is, at best, equal to that of the larger Transformer. However, scaling up model hyperparameters does not seem to solely resolve all the problems. Fig. \ref{fid} shows that T5 performs significantly worse than some open-domain QA models despite substantially larger computes spent for the training. Likewise, \citet{marge} shows that MARGE outperforms various baselines that consume substantially more computes and use more task-specific human intervention. Thus, we may not only need conditional computation but also retrieval-based approaches in order to improve the performance-computes trade-off further. 

\end{document}

%% file: math_commands.tex
\usepackage{amsmath,amsfonts,bm}

\def\eqref#1{equation~\ref{#1}}

\def\1{\bm{1}}

\DeclareMathAlphabet{\mathsfit}{\encodingdefault}{\sfdefault}{m}{sl}
\SetMathAlphabet{\mathsfit}{bold}{\encodingdefault}{\sfdefault}{bx}{n}

